\documentclass[10pt, conference]{IEEEtran}

\usepackage{array,multirow}

\usepackage{longtable}
\usepackage[style=numeric, citestyle=numeric-comp, maxnames=3, giveninits, sorting=none, natbib, backend=bibtex]{biblatex}

\usepackage[symbol]{footmisc}
\usepackage{algorithm}
\usepackage{algorithmic}
\usepackage{amsmath} 
\usepackage{hyperref}
\usepackage{booktabs}

\ifCLASSINFOpdf
    \usepackage{subfig}
    \usepackage{caption}
    \usepackage{graphicx}
    \usepackage{svg}

\else

\fi

\begin{document}

\title{An Adaptive Indoor Localization Approach Using WiFi RSSI Fingerprinting with SLAM-Enabled Robotic Platform and Deep Neural Networks} 

\author{\IEEEauthorblockN{Seyed Alireza Rahimi Azghadi\IEEEauthorrefmark{1}, Atah Nuh Mih\IEEEauthorrefmark{1}, Asfia Kawnine\IEEEauthorrefmark{1}, Monica Wachowicz\IEEEauthorrefmark{1}\IEEEauthorrefmark{2}, Francis Palma\IEEEauthorrefmark{1}, Hung Cao\IEEEauthorrefmark{1}}

\IEEEauthorblockA{\IEEEauthorrefmark{1} \textit{Analytics Everywhere Lab, University of New Brunswick, Canada} \\
 \IEEEauthorrefmark{2} \textit{RMIT University, Australia} \\
% \IEEEauthorrefmark{3} \textit{RMIT University, Australia} \\
% \IEEEauthorrefmark{4} \textit{Department of Mechanical Engineering, University of New Brunswick, Canada } \\
}
}

\maketitle

\begin{abstract}
Indoor localization plays a vital role in the era of the Internet of Things (IoT) and robotics, with Wireless Fidelity (WiFi) technology being a prominent choice due to its ubiquity. We present a method for creating WiFi fingerprinting datasets to enhance indoor localization systems and address the gap in WiFi fingerprinting dataset creation. We used the Simultaneous Localization And Mapping (SLAM) algorithm and employed a robotic platform to construct precise maps and localize robots in indoor environments. We developed software applications to facilitate data acquisition, fingerprinting dataset collection, and accurate ground truth map building. Subsequently, we aligned the spatial information generated via the SLAM with the WiFi scans to create a comprehensive WiFi fingerprinting dataset. The created dataset was used to train a deep neural network (DNN) for indoor localization, which can prove the usefulness of grid density. We conducted experimental validation within our office environment to demonstrate the proposed method's effectiveness, including a heatmap from the dataset showcasing the spatial distribution of WiFi signal strengths for the testing access points placed within the environment. Notably, our method offers distinct advantages over existing approaches as it eliminates the need for a predefined map of the environment, requires no preparatory steps, lessens human intervention, creates a denser fingerprinting dataset, and reduces the WiFi fingerprinting dataset creation time. Our method achieves 26\% more accurate localization than the other methods and can create a six times denser fingerprinting dataset in one-third of the time compared to the traditional method. In summary, using WiFi RSSI Fingerprinting data surveyed by the SLAM-Enabled Robotic Platform, we can adapt our trained DNN model to indoor localization in any dynamic environment and enhance its scalability and applicability in real-world scenarios.
\end{abstract}


\IEEEpeerreviewmaketitle

\section{Introduction}
\label{sec:introduction}

Indoor localization has become a crucial aspect of our lives. With the advent of the Internet of Things (IoT), more devices are connected, and the demand for location-based services is increasing \cite{basiri_indoor_2017, cao2020holistic, cao2023fostering}. Indoor localization involves determining the precise location of objects or individuals within a building or other indoor environments \cite{turgut_indoor_2016}. Various technologies have been employed to meet the growing demand for accurate indoor localization, each with advantages and limitations. Wireless Fidelity (WiFi) \cite{estrada_wifi_2023, sarcevic_indoor_2023, molina_multimodal_2018}, Ultra-wideband (UWB) \cite{yu_novel_2019}, Bluetooth Low Energy (BLE) \cite{phutcharoen_accuracy_2020}, and UltraSonic \cite{carotenuto_indoor_2019} technologies have been considerably utilized to provide precise indoor positioning. However, each technology and method has its strengths and limitations, WiFi-based solutions stand out among these technologies. WiFi networks are prevalent in indoor environments, making them easily accessible and cost-effective for deploying location-based services \cite{shang_overview_2022}, so WiFi-based localization is a preferred choice for indoor localization in many applications. 

\begin{figure}[t!]
  \centering
  %\includesvg[width=1\linewidth]{figs/introduction_chart.svg}
  \includegraphics[width=1\linewidth]{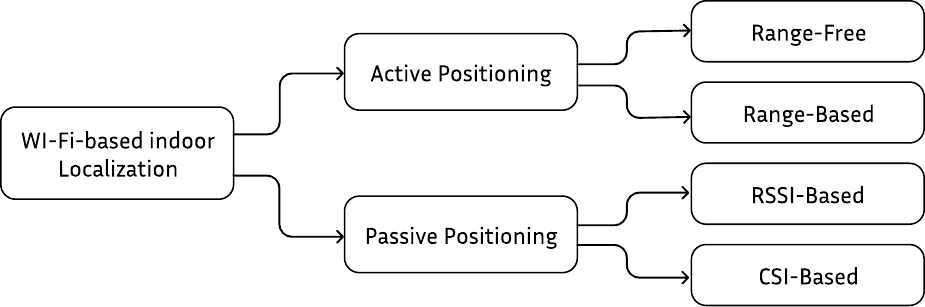}
  \caption{Classification of WiFi-based Indoor Localization}
  \label{hierarchical_methods}
  \vspace{-0.75cm}
\end{figure}

WiFi technology has been harnessed for indoor localization through various methods, classified into active and passive approaches based on the presence of a WiFi module. Active indoor localization involves the active participation of a device equipped with WiFi modules. In contrast, passive localization relies on monitoring existing WiFi signals without requiring active participation from the object \cite{liu_survey_2020}. Within the Active category, algorithms can be classified as Range-Based or Range-Free. Range-Based algorithms rely on signal strength measurements, time of flight, or angle of arrival to estimate the distance between the device and WiFi access points \cite{liu_survey_2020, shang_overview_2022, dai_survey_2023}. These methods often require additional hardware and infrastructure for precise localization \cite{mesmoudi_wireless_2013}. On the other hand, Range-Free algorithms eliminate the need for distance measurements and instead use connectivity patterns or signal presence for positioning. Instead, they necessitate an offline step to create a fingerprinting dataset, which involves mapping the WiFi signals in the environment. Figure \ref{hierarchical_methods} demonstrates the hierarchical relation of different methods utilized in WiFi-based indoor localization \cite{liu_survey_2020}.

WiFi fingerprinting, a method of associating signal characteristics with indoor locations, allows for accurate indoor positioning without relying on line-of-sight (LoS) assumptions or precise distance measurements. This approach offers robustness against multipath effects and non-LoS propagation, making it highly feasible and advantageous for indoor localization \cite{shang_overview_2022}. However, the current fingerprinting-based WiFi localization methods have made significant strides \cite{estrada_wifi_2023, sarcevic_indoor_2023, rana2023indoor, molina_multimodal_2018}, a research gap exists in the process of creating WiFi fingerprinting datasets, which is a crucial element for the effectiveness of algorithms. 

Fingerprinting localization methods usually have two phases: Online and Offline. During the Offline phase, we should create a map from the environment based on the Received Signal Strength Indicator (RSSI) or any other feasible technology. Then, this map is used to train a localization model that can estimate the location based on the map's information. While in the Online phase, a device can collect the same information from the environment without knowing the exact location, so with the help of the localization model, it can localize itself within the map.

As existing WiFi fingerprinting localization methods require an offline step for dataset creation, there is room for enhancement of fingerprinting dataset collection methods in various aspects. Many recent papers in this field use on-ground griding for collecting WiFi signal strengths to make their datasets, which takes a lot of time and effort \cite{estrada_wifi_2023, sarcevic_indoor_2023}. In addition, by traditionally collecting WiFi signals, the density of the sample points will be small, resulting in poor location prediction accuracy. Another issue occurs if the surveyed environments change, which makes the localization model and fingerprinting dataset invalid. Updating the fingerprinting dataset will be time-consuming if the data collection approach is not fast and efficient.

Therefore, we propose a novel approach leveraging a commercialized robotic platform to streamline and simplify the dataset creation process. Our method aims to overcome the limitations of current approaches, providing more efficiency and flexibility. We developed a Python controller program based on Robot Operating System 2 (ROS2) \cite{ros2_2022} API that lets us control our mobile Robot and used a 2D Light Detection And Ranging (LiDAR) scanner with the help of the Simultaneous Localization And Mapping (SLAM)\_TOOLBOX \cite{macenski_slam_2021} algorithm to localize the Robot in the environments. Based on this accurate auxiliary localization method, we could collect a dense fingerprinting dataset of RSSI, which lets us train a localization model that provides more accuracy than state-of-the-art methods. Our main contributions in this study include:
\begin{itemize}
    \item \textbf{A Novel Dataset Creation Method}: Using a robotic platform, we simplified the process of creating indoor localization datasets, making them more efficient and flexible.
    \item \textbf{Improvements to Localization Accuracy with Deep Neural Networks}: We trained a deep neural network that significantly improves indoor localization accuracy for real-world use.
    \item \textbf{A Study on the Influence of Reference Point Density on Model Performance}: We studied how the number of reference points (RPs) affects the accuracy of a DNN localization model.
\end{itemize}

The paper is structured as follows: \autoref{sec:related} presents related work considering different approaches for WiFi fingerprinting dataset creation for indoor localization and WiFi localization methods, and we discuss their methodologies, strengths, and limitations. In \autoref{sec:method}, we explained the steps of our approach for creating a WiFi fingerprinting dataset followed by \autoref{sec:implementation}, which present implementation details for each step of the proposed method. \autoref{sec:experiement} presents information about our experiment, showing our approach's practicality. In the \autoref{sec:evaluation}, we compared our approach with related works on time efficiency, adaptability, and accuracy. Finally, we presented possible ideas for future work and summarized our contribution and work in the \autoref{sec:conclusion}.

\section{Related Work}
\label{sec:related}

Several researchers have addressed the challenges of indoor localization using various sensors and algorithms to enhance accuracy and flexibility. Here, we provide an overview of relevant studies. We highlighted their methodologies, contributions, and identified their shortcomings. Section \ref{sub1} discusses studies that contributed to improving fingerprinting-based datasets in different aspects, and Section \ref{sub2} summarizes studies that focused on building more accurate localization models.

\subsection{Fingerprinting Dataset Collection}\label{sub1}
Rizk et al. \cite{rizk_laser_2023} used a 2D LiDAR scanner for individual tracking inside a room to collect WiFi fingerprinting samples. The method is mainly based on individuals walking around the room with mobile devices that can measure RSSI and record the information. The method's main advantage over traditional fingerprinting is its use of LiDAR, which speeds up the process and builds the map accurately. Nevertheless, there is still room to speed up the collection process. In big or complex environments, the approach will have problems as the authors have to relocate the LiDAR sensor due to its limited range or out-of-sight places. The sensor relocation will cause considerable overhead to the system as the placement locations must be measured, and the collected data must be aligned. Also, the proposed method did not address the challenge of building a dense map. In addition, the approach relies on individuals roaming around the room and gathering the WiFi RSSI information; therefore, updating and creating the dataset requires multiple individuals, which reduces efficiency and flexibility. Abu Kharmeh et al. \cite{abu_kharmeh_indoor_2023} introduced a robot-driven dataset construction framework. The authors used a custom robot to follow black tape on the ground to build a multi-height WiFi fingerprinting dataset. The proposed dataset contributes to the indoor localization systems by providing multi-level WiFi RSSI. However, the data collocation methodology has multiple limitations: (1) the authors did not build dense maps with tight reference points (RPs) because their Robot follows a black tape grid on the ground to collect data and gather data only at specific cross points, and (2) the robot navigation system causes inflexibility and requires significant manual labor. 

Silva et al. \cite{silva_industrial_2023} accumulated a WiFi RSSI dataset using monitoring devices installed on a manually pushed trolley within an industrial setting. The authors' main contribution is a public dataset gathered from multiple sensors in an industrial setting, which can help indoor tracking solutions with localization systems. However, their method can not build a dense map due to the localization difficulty for manually pushed trolly. In addition, the authors used computer vision to build ground truth, although they did not provide many reference points due to the overhead of installing ArUco tags. Thus, collecting data would be time-consuming and inefficient. Abdullah et al. \cite{abdullah_utmindualsymfi_2023} developed a Windows program with Matlab for Dual-band WiFi RSSI collection. The authors have collected WiFi samples from different floors of four buildings with two different laptops. Also, the authors have provided an extensive analysis of their dataset that is their main contribution. However, they have failed to collect enough RPs. As they manually measured the distance between the RPs, collecting many RPs would be time-consuming and difficult. 

\subsection{Localization Methods}\label{sub2}
Molina et al. \cite{molina_multimodal_2018} proposed an experimental work at a university and airport that uses a weighted KNN algorithm. The authors built a localization system with data that they had gathered from Access Points (APs) and Ibeacons. In the proposed method, the authors used WiFi and BLE modules to collect samples for building an offline fingerprinting dataset that is later used to determine the users' location. They have merged multimodal signals to build an accurate localization model, although there is still room to enhance the prediction system. The proposed system's algorithm and sparsity of the collected RPs lead to high localization errors because they used traditional data collection methods that are not efficient. Sarcevic et al. \cite{sarcevic_indoor_2023} developed a novel approach to estimate robots' location in a 2D space. Unlike other works that only use communication protocols' signal strength or features, the authors integrated the magnetometer data of the building to increase their accuracy. The authors evaluated the proposed method in two scenarios and achieved good results. Although the proposed approach has some drawbacks. First, magnetometer readings can change easily at different heights due to the absence of close enough building structures, as these structures can affect the earth's magnetic field. Besides, the designed experiment differs from real-world situations because the authors have placed multiple APs in the experiments' rooms in line-of-sight. Finally, using a grid to divide the area into multiple points for gathering data is not promising. It will create a low-density dataset and cause manual labor, resulting in inefficiency.

Rana et al. \cite{rana2023indoor} used a relatively newer technology in WiFi AP, which provides the round trip time (RTT) instead of the RSSI. Like the other fingerprinting methods, the authors used the gathered information to train a localization model that is based on a Deep Neural Network (DNN) and a Random Forest (RF) to estimate the location in 2D. The study's experiment is evaluated with the collected data by a smartphone from a room with a 1m grid that is a huge gap compared to the other methods. The authors have contributed to the indoor localization system by developing a new localization model and integrating new technology for fingerprint creation. However, the mentioned RTT feature is only available in relatively new infrastructure. Also, Updating and maintaining the APs inside an institution does not happen often, so the whole data type does not seem practical. In addition, the authors could have used a more efficient data collection method and built a more dense fingerprinting dataset with more RPs other than the one-meter grid.

In summary, while these studies contribute significantly to the field, they exhibit common shortcomings, including the inability to construct dense maps, which will result in low-accuracy localization. In addition, the essential role of the individuals in the data collection process prevents efficiency. Finally, with manual data labeling for the ground truth of the collected data sample locations, the dataset creation process would be time-consuming. These limitations motivate the development of a novel technique, as proposed in our research, addressing these gaps and advancing the state-of-the-art in fingerprinting dataset construction with a mobile robot.

\begin{figure}[t!]
  \centering
    %\includesvg[width=1\linewidth]{figs/method_overview.svg}
    % \includegraphics[width=.7\textwidth]{figs/workflow.png}
    \includegraphics[width=1\linewidth]{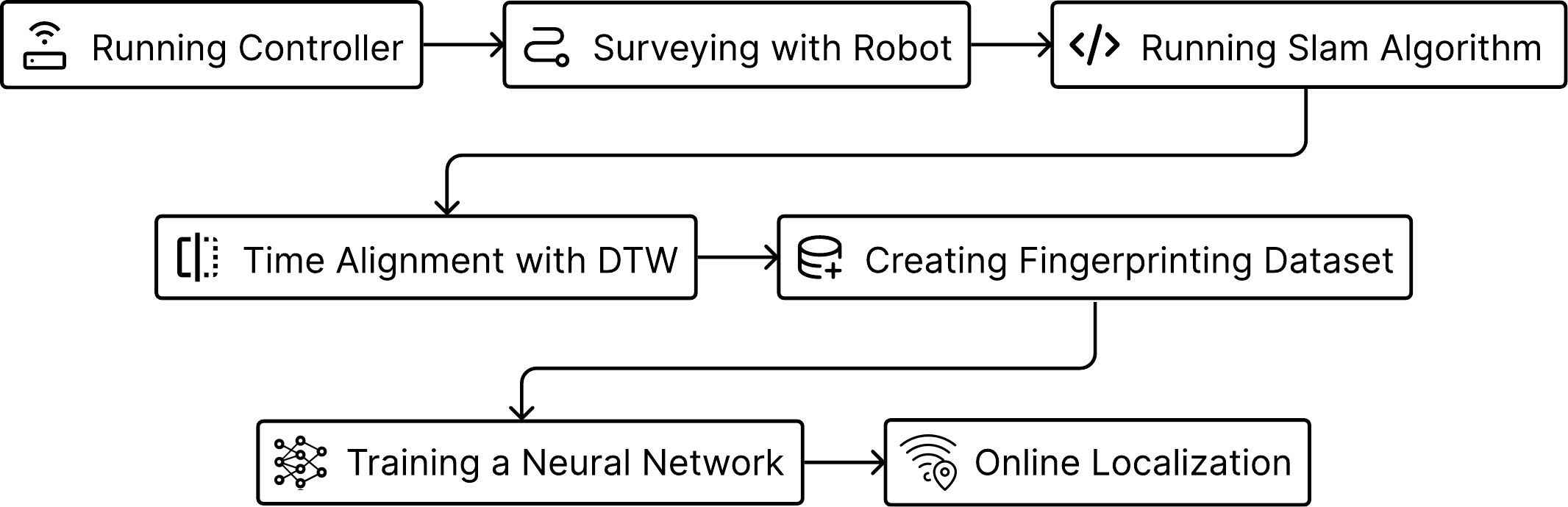}
  \caption{Method Overview}
  \label{fig:method_overview}
\end{figure}

\section{Proposed Method}
\label{sec:method}

This section briefly overviews our proposed method, and Figure \ref{fig:method_overview} shows the steps we have taken to create the fingerprinting dataset and localization. Our method involves creating a detailed dataset of WiFi RSSI in an indoor environment using a robotic platform and training a deep neural network on this dataset to enable online indoor localization based solely on WiFi signal strength. The dataset creation process begins with Python3 controller execution on the Robot, which interacts with the ROS2 API, allowing us to control the Robot through a web interface as it moves around the environment. We navigate the Robot to collect WiFi data while mapping its surroundings using a LiDAR sensor with the SLAM\_TOOLBOX. SLAM creates a 2D map and provides the Robot's exact location on the map in real-time.

After data collection, we synchronize the WiFi RSSI with the Robot's position over time using Dynamic Time Warping (DTW) because the Robot's position and WiFi samples do not have the same sampling rate, and their sequences might not start from the same timestamp. Finally, we convert all the collected data into CSV format, providing a clear overview of the Robot's position over time and the corresponding WiFi RSSI. With this dataset, we train a neural network to learn the hidden relationships between the WiFi samples and their corresponding locations. This model can be used for the online phase of indoor localization and different applications. As embedded devices become more powerful, trained neural networks can be deployed on mobile phones for localization and navigation inside complex buildings such as universities or malls.

\begin{figure*}[t]
  \centering
    %\includesvg[width=\linewidth]{figs/rosgraph.svg}
    \includegraphics[width=1\linewidth]{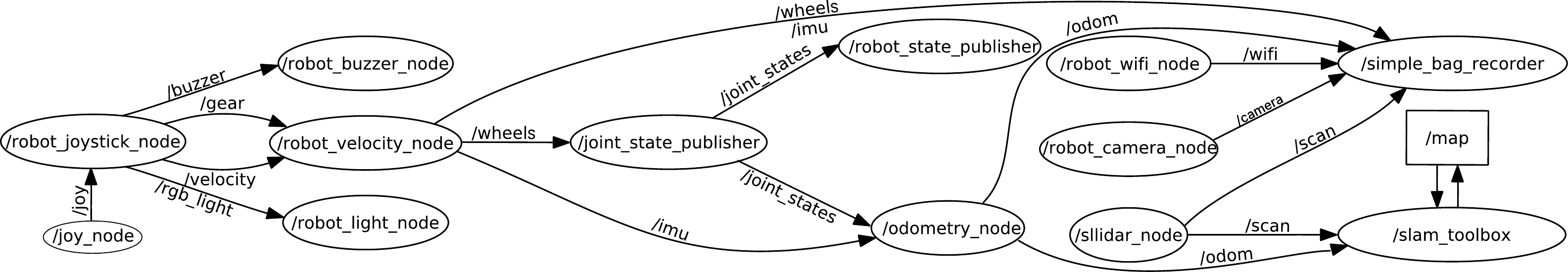}
  \caption{ROS2 topics and nodes interaction in the robot}
  \label{fig:ros2_fig}
\end{figure*}

\section{Implementation}
\label{sec:implementation}

The experiment involves integrating a commercialized robotic platform, developing a comprehensive methodology for data collection and map building, and, finally, utilizing a deep neural network for indoor localization. This section provides a detailed description of each step employed in this research.

\subsection{\textbf{Running Controller}}

The first step involves executing Python controller code on the Robot, enabling interaction via a web interface developed using FASTAPI through the Robot's WiFi hotspot. This interface facilitates the initiation and termination of surveys and the retrieval of recorded survey information. The Python program was developed to manage the interaction with sensors and actuators, recording all events, messages, and sensor readings on various ROS2 topics. The application enhances the flexibility of experimentation and streamlines data collection processes. This program harnesses the capabilities of the ROS2 API, enabling seamless communication and control of various robotic elements.

Additionally, the asynchronous architecture of ROS2 allows multiple nodes to operate in parallel, with each node responsible for a specific task. We developed multiple nodes for tasks such as velocity and movement commands, joystick integration, LiDAR sensor, IMU sensor, camera, and calculating odometry based on wheel encoders. Nodes may contain drivers that enable low-level hardware interaction with specific sensors or actuators. Figure \ref{fig:ros2_fig} illustrates the interaction of different elements of the robotic system through ROS2 topics generated by the RQT program, which is available in ROS2 libraries when the developed application is executed on the robotic platform. This graphical representation encapsulates the flow of information, providing insights into the intricate communication channels within the robotic ecosystem. In optimizing the WiFi setup for seamless data collection within the robotic system, specific considerations were made regarding channel selection. Utilizing channels 1, 6, and 11 was a strategic choice to minimize WiFi's scan gap and maximize efficiency. These channels were chosen due to their non-overlapping nature, reducing signal interference likelihood. Moreover, these channels are commonly configured on APs, ensuring compatibility and enhancing communication reliability. As a result of scanning fewer channels during the same period, we can scan more during the surveying period and collect more WiFi signal samples from the environment.

\subsection{\textbf{Surveying With Robot}}

\begin{figure}[t]
  \centering
    \includegraphics[width=\linewidth]{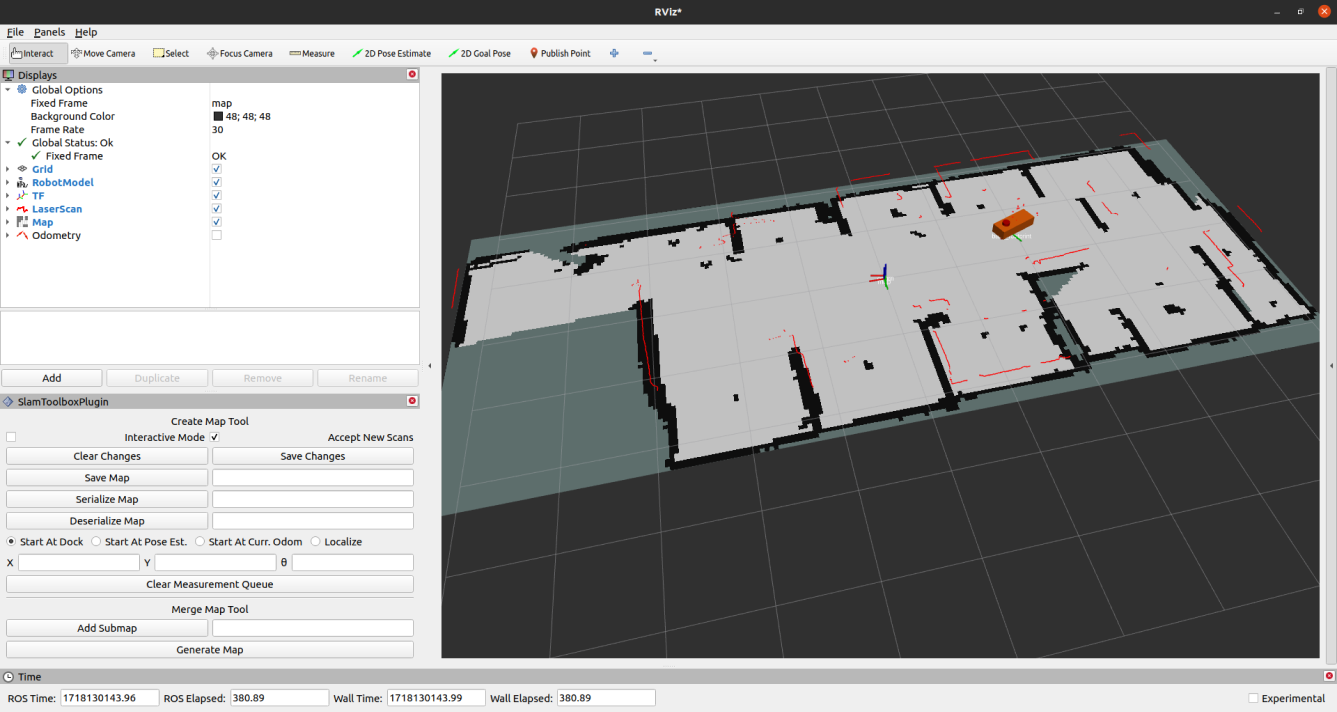}
  \caption{RVIZ ROS2 experiment monitoring during the test}
  \label{fig:rviz}
\end{figure}

Following the execution of the controller, the Robot is positioned in the center of the room, and environmental surveys are conducted. Utilizing joystick controls, the Robot navigates through the environment, systematically covering different areas to capture comprehensive WiFi information. All sensor inputs are stored in a ROS bag file for subsequent analysis and future reference. By defining the ROS\_DOMAIN variable on both the Robot and the development systems, we can monitor live information and messages transferred among different ROS2 topics through the WiFi network. Figure \ref{fig:rviz} shows the development environment during the surveying process, which aids debugging. On the left side, various ROS2 topics are available for monitoring, and on the right side, the Robot's position relative to the generated map is displayed. We defined our robot parameters and description precisely in a URDF file, an input file for the ROS2 Robot State Publisher Node. The graphical representation of the Robot on the map is accurate and matches the actual Robot's size and dimension, generated by the mentioned node. Precise robot definition plays an essential role in the accuracy of the SLAM method, as map creation and localization depend on sensors' positions, sensors' readings, and odometry.

\subsection{\textbf{Running SLAM Algorithm}}

\begin{figure*}[t!]
  \centering
    \subfloat[\centering RSSI Heatmap]{{%\includesvg[width=0.45\linewidth]{figs/heatmap.svg} 
    \includegraphics[width=0.45\linewidth]{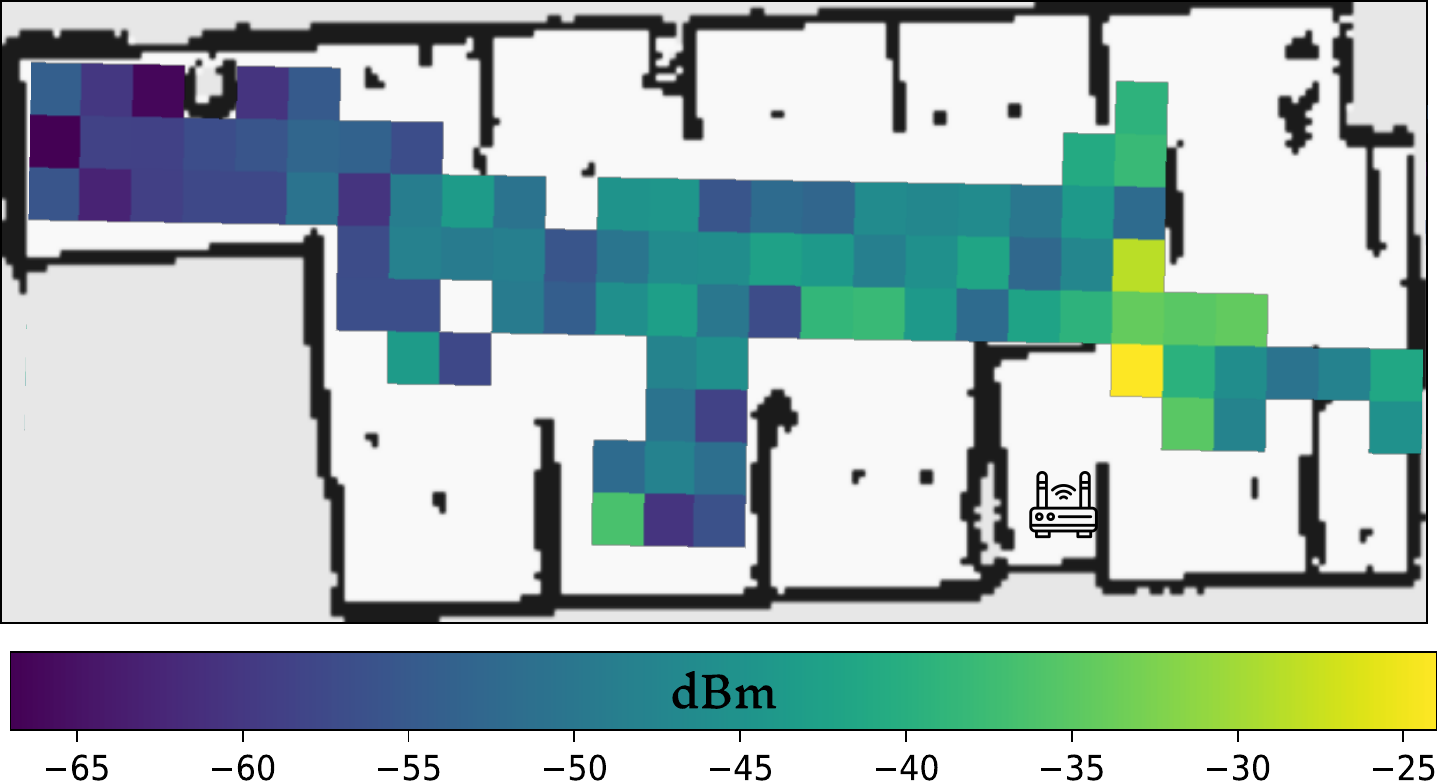}
    }}
    \qquad
    \subfloat[\centering Location Samples on Map]{{%\includesvg[width=0.45\linewidth]{figs/robot_locations.svg}
    \includegraphics[width=0.45\linewidth]{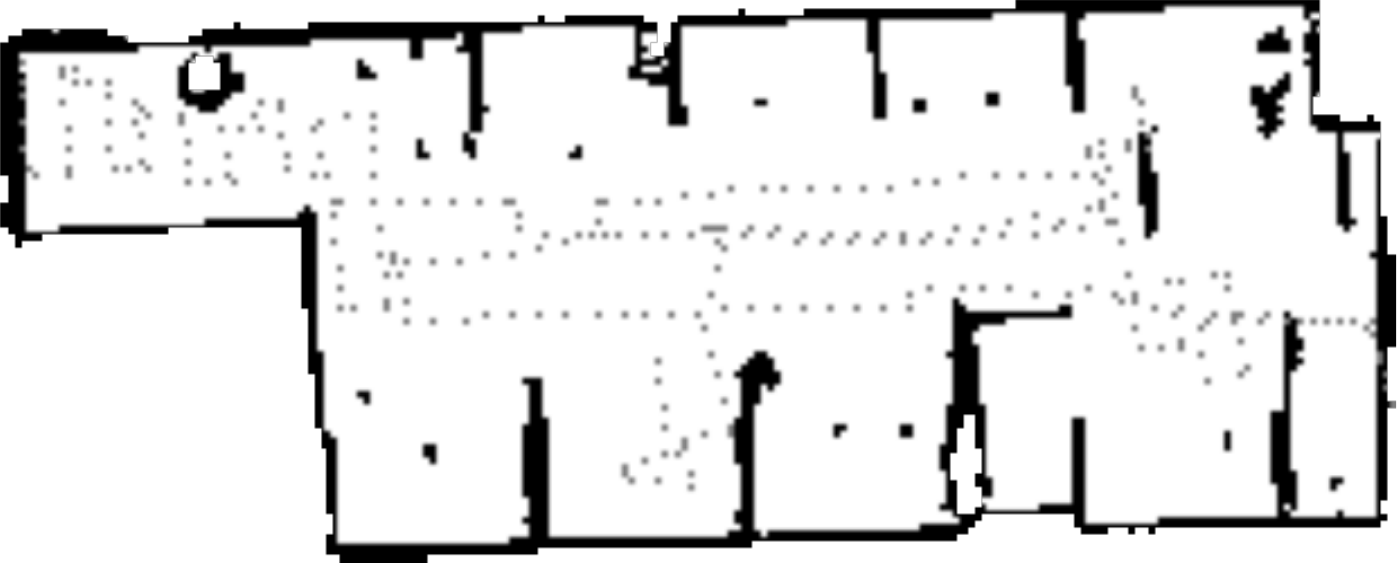}
    }}
  \caption{Fingerprinting Dataset Visualization}
  \label{fig:fingerprint}
\end{figure*}

The SLAM\_TOOLBOX, chosen for its accuracy and open-source nature, is employed with a single 2D LiDAR sensor. The toolbox serves as a ROS2 node that subscribes to the Odom topic and publishes on the transform and map topics. Our Python application provides the Odom topic, which calculates the Robot's position over time based on sensor values. In our experiments, we used wheel encoders and an IMU sensor to calculate the odometry. Although the provided odometry information based on the sensors is accurate, the estimated location contains some errors over time due to robot drift, IMU sensor accumulative error, and random noise. The SLAM\_TOOLBOX enhances the odometry system by using scan matching as the Robot moves through the environment. As we survey the environment, a 2D map is built and published on the map topic by SLAM\_TOOLBOX. Based on the generated map up to the current time and a new LiDAR scan sample, SLAM\_TOOLBOX can estimate the Robot's location within the map. Additionally, since the initial location of the Robot inside the map is known, a transformation from odometry to the map is published at each timestamp, showing the exact amount of odometry drifts up to the current time. This whole correction process results in a more accurate WiFi fingerprinting map.

\subsection{\textbf{Time Alignment with DTW}}

Direct pairwise matching between WiFi scans and robot odometry is challenging because the SLAM-generated trajectory has a different publishing rate than the WiFi scans. The frequency of WiFi scan data recorded in ROS2's bag is lower (1Hz) than that of the Robot's location data (100Hz) published by the Odom and SLAM algorithms. In addition, the starting time for saving the odometry samples and WiFi scans is different. To address these issues, DTW is employed to align the two time series, creating a pairwise match between each scan and its nearest location with the smallest error. For clarity and better understanding, we provide the pseudo-code of the DTW in Algorithm \ref{algo:dtw}. DTW is a technique used to measure the similarity between two temporal sequences by aligning them through stretching or compressing time, allowing for non-linear correspondence despite variations in timing and speed. It is particularly useful for comparing sequences with varying lengths or time scales, such as robot odometry data and WiFi scans.

\begin{algorithm}
\caption{Dynamic Time Warping}
\begin{algorithmic}[1]
\REQUIRE Sequence \( A = \{a_1, a_2, \ldots, a_n\} \) 
\REQUIRE Sequence \( B = \{b_1, b_2, \ldots, b_m\} \)
\STATE Initialize a cost matrix \( C \) of size \( (n+1) \times (m+1) \) with all elements as \( \infty \)
\STATE Set \( C[0][0] = 0 \)
\FOR{\( i = 1 \) to \( n \)}
    \FOR{\( j = 1 \) to \( m \)}
        \STATE Compute the local cost \( d(a_i, b_j) \)
        \STATE \( \text{cost} = d(a_i, b_j) + \min(C[i-1][j], C[i][j-1], C[i-1][j-1]) \)
        \STATE \( C[i][j] = \text{cost} \)
    \ENDFOR
\ENDFOR
\RETURN \( C[n][m] \)
\end{algorithmic}
\label{algo:dtw}
\end{algorithm}

\subsection{\textbf{Creating Fingerprinting Dataset}}
We constructed a fingerprinting dataset with the aligned data detailing the Robot's position over time and the associated WiFi signal strengths from identified APs. This fingerprinting dataset can be used to construct a heatmap, such as Figure \ref{fig:fingerprint}, which represents spatial variations in WiFi signal strengths, or to train a model for localization. Figure \ref{fig:fingerprint} (a) shows the signal strength of an AP placed inside the lab for the experiment(the exact location of the access point is shown in the figure). The heatmap is built by calculating the average RSSI of different samples within each grid box and plotting the average value using the Viridis color map. Places not covered with the heatmap do not have any data samples because of the obstacles, such as the chairs and tables, that hindered our Robot from navigating there. A guideline for understanding the signal strengths is provided at the bottom of the figure to aid interpretation. Stronger signals are illustrated with brighter colors, while weaker signals are shown with darker colors. At the bottom center of the image and the top left corner, unusually powerful signals are observed, probably caused by signal reflections from obstacles. Figure \ref{fig:fingerprint} (b) shows the aligned locations of the samples where we have WiFi scans. This dataset, organized in CSV format, contains rows indicating the Robot's positions over time and columns representing RSSI values from all detected WiFi APs throughout the survey. It is common for some access points to be present only in certain parts of the map while being out of reach in other areas. As we built the fingerprinting dataset, we assigned null values for each access point that is not available at a particular location, making it easier to handle in future steps. We provide a sample output of our method in Table \ref{tab:WiFi-dataset}. The resulting CSV file serves as the final output of the process.

\begin{table}[t!]
\centering
\caption{WiFi Fingerprinting Dataset Format}
\begin{tabular}{c c c c c c}
\toprule
Timestamp & $X\_Pos$ & $Y\_Pos$ & $MAC_1$ & $\cdots$ & $MAC_N$ \\
\midrule
1707935831.6001 & 0.0000 & 0.0000 & 66.0 & $\cdots$ & NaN \\
1707935832.5993 & 0.0026 & 0.0034 & 66.0 & $\cdots$ & NaN \\ 
$\cdots$ & $\cdots$ & $\cdots$ & $\cdots$ & $\cdots$ & $\cdots$ \\
1707935963.4922 & 3.4581 & 10.101 & 70.0 & $\cdots$ & 83.0 \\
\bottomrule
\end{tabular}
\vspace{0.2cm}

\label{tab:WiFi-dataset}
\end{table}

\subsection{\textbf{Training a Neural Network}}

\begin{figure}[t!]
  \centering
    %\includesvg[width=0.7\linewidth]{figs/neural_network.svg}
    \includegraphics[width=0.7\linewidth]{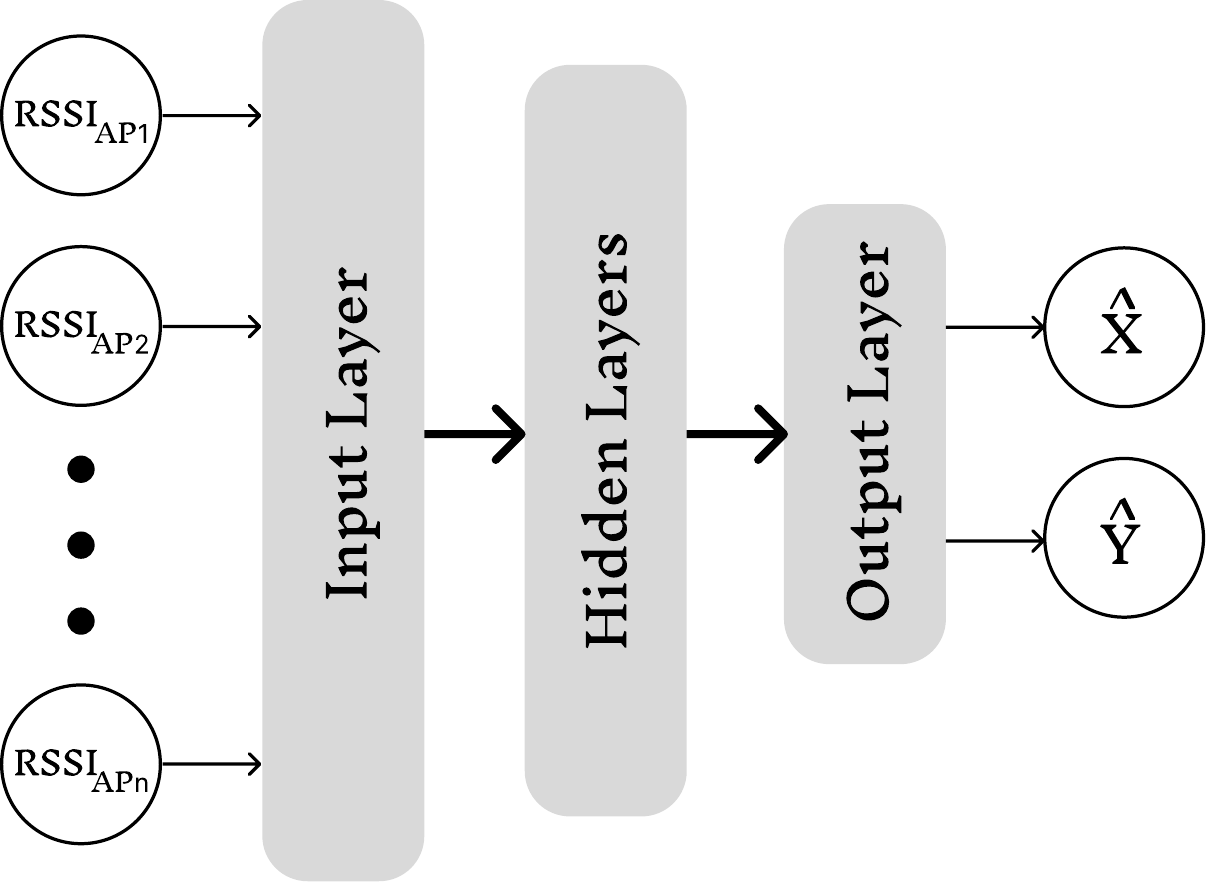}
  \caption{Neural Network Architecture}
  \label{fig:network}
\end{figure}

\begin{figure}[t!]
  \centering
    %\includesvg[width=1\linewidth]{figs/loss.svg}
    \includegraphics[width=1\linewidth]{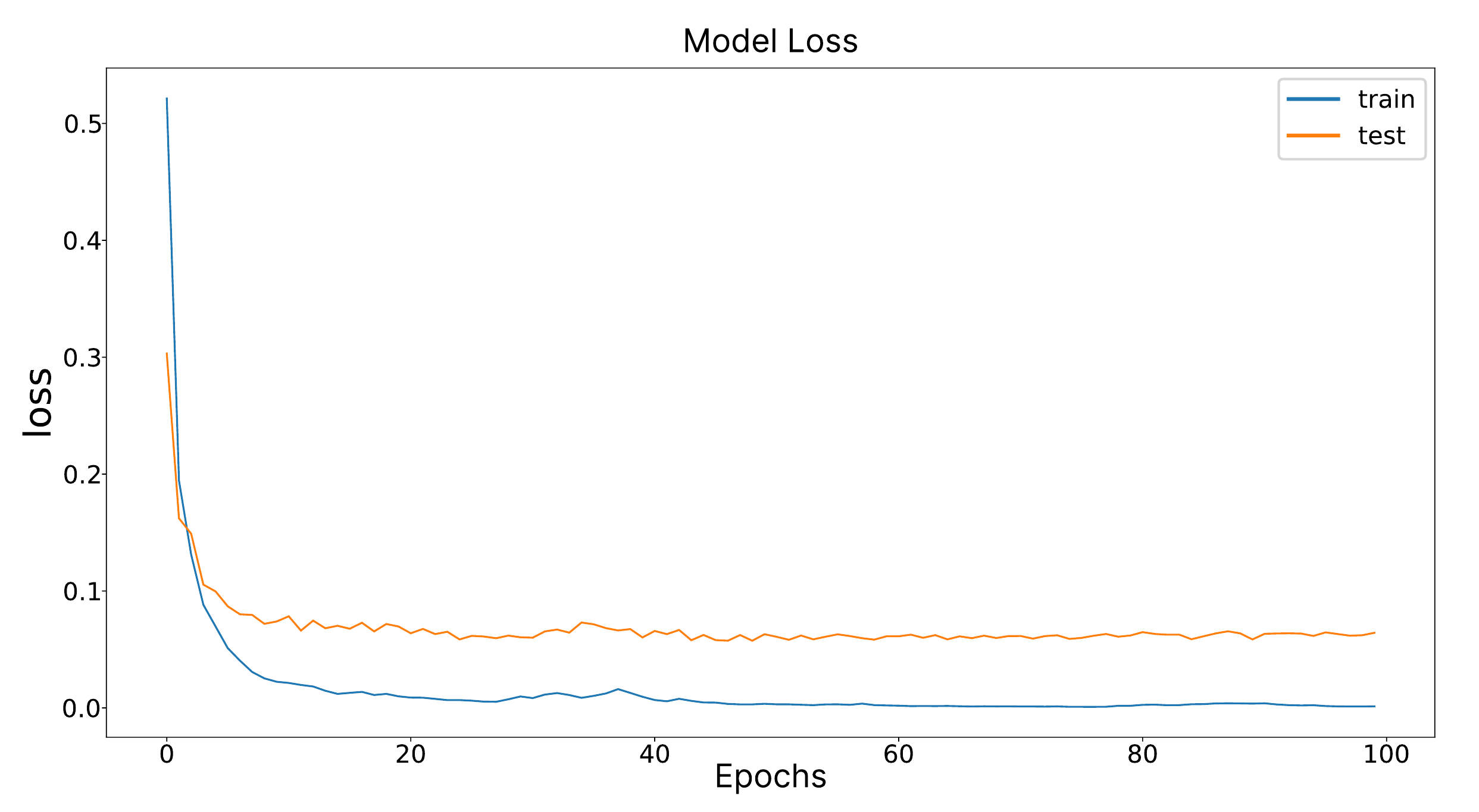}
  \caption{Model Training Loss}
  \label{fig:loss}
\end{figure}

After creating a fingerprinting dataset, we can train a model for indoor localization. The model can predict the location where we collect signal strength samples. We designed a deep neural network architecture for doing the mentioned task, which gets a list of signal strengths for an RP and can predict the $x$ and $y$ for the given point. The architecture of the trained network is illustrated in Figure \ref{fig:network}. The input layer size is equal to the total number of access points we found in any experiment, and the output layer dimension is 2 for the prediction of the width and height of the sample from the map origin. A description of the neural network architecture is provided in Table \ref{tab:dnn}. The network has four layers with the mentioned dimensions, and the activation function of the first three layers is \textit{ReLU}. However, to predict the location, which can also be negative as it is relative to the origin, we used a \textit{linear} activation function on the output layer. We used \textit{Adam} optimizer and trained the network for 100 epochs with \textit{Early Stopping} to prevent overfiting, and our loss function is the Mean Absolute Error (MAE) function. Figure \ref{fig:loss} is the model loss during the training phase, which shows the model convergence and provides information that our model did not overfit. In addition, as the neural networks cannot tolerate missing values, for the training, we replaced the null values for the out-of-reach access points with -100dB, which represents weak signal strength.

\begin{table}[t!]
    \centering
    \caption{Neural Network Description}
    \begin{tabular}{c c}
        \toprule
        \textbf{Parameter} & \textbf{Description} \\
        \hline
        Input Layer & Signal Strength of all APs \\
        Output Layer & Predicts a 2D location \\
        Layer Sizes & 256, 128, 32, 2 \\
        Activations & ReLU, ReLU, ReLU, Linear \\
        Loss Function & Mean Squared Error \\
        Optimizer & Adam \\
        Mini-batch size & 32 \\
        Early Stopping Patience & 5 \\
        \bottomrule
    \end{tabular}
    \label{tab:dnn}
\end{table}

\subsection{\textbf{Online Localization}}
After training a neural network in the online phase, we can use our model to predict a location based on the live WiFi signal strength. We developed a deep neural network with a few parameters; this network can easily be deployed on mobile devices such as users' phones or other IoT devices, which might require indoor localization. Different factors, such as device heterogeneity, might affect the model performance, but we can still localize the user within the map in real-time. At any time that localization is demanded, we can scan WiFi signals available in the environment, and with an inference from the model, we can estimate the position. Nevertheless, there are more steps to make this research even more practical; we can survey a building and develop an application for users' phones to make indoor localization easily accessible to the end users.

\section{Experiments}
\label{sec:experiement}
This section explains the various parts of our experiment, including (1) the platform and the equipment that we used for our real-world experiment, (2) the experiment location, and (3) the ground truth. We used a commercialized robotic platform to increase the method's adaptability and practicality. We tested our method inside our office, simulated a real-world scenario, and presented our approach in the explained environment. We collected ground truth data and compared the signal strength to prove the validity of our fingerprinting dataset collection approach.

\subsection{\textbf{Robotic Platform}}
The experimental framework depends on Yahboom's Rosmaster X3 \cite{yahboom}, a commercially available robotic platform, as the physical basis for data gathering and localization. The Rosmaster X3 platform serves as the hardware foundation, providing the necessary sensor and actuator integration interfaces. Our robotic platform is equipped with multiple sensors such as 2D LiDAR, WiFi module, Cameras, IMU, and wheel encoder, which we provided extra information about each sensor in Table \ref{tab:robot_info}. Our Robot's core element is a Jetson Orin Nano, which serves as our robot brain. There are two other auxiliary boards connected to the Jetson. First, a robotic expansion board powered by an STM32 microcontroller, which interacts with analog devices such as IMU and wheels. Second, another board is a USB hub, which provides enough power and ports for other sensors installed on the Robot. Jetson runs Ubuntu 20.04 as its operating system, and the ROS2 Galactic version is installed. However, Jetson is equipped with a powerful Arm architecture processor with six cores and 4GB of RAM; it is not an ideal system for development. Thus, our system is developed and debugged on a desktop. 

\begin{figure}[t!]
  \centering
    \includegraphics[width=0.8\linewidth]{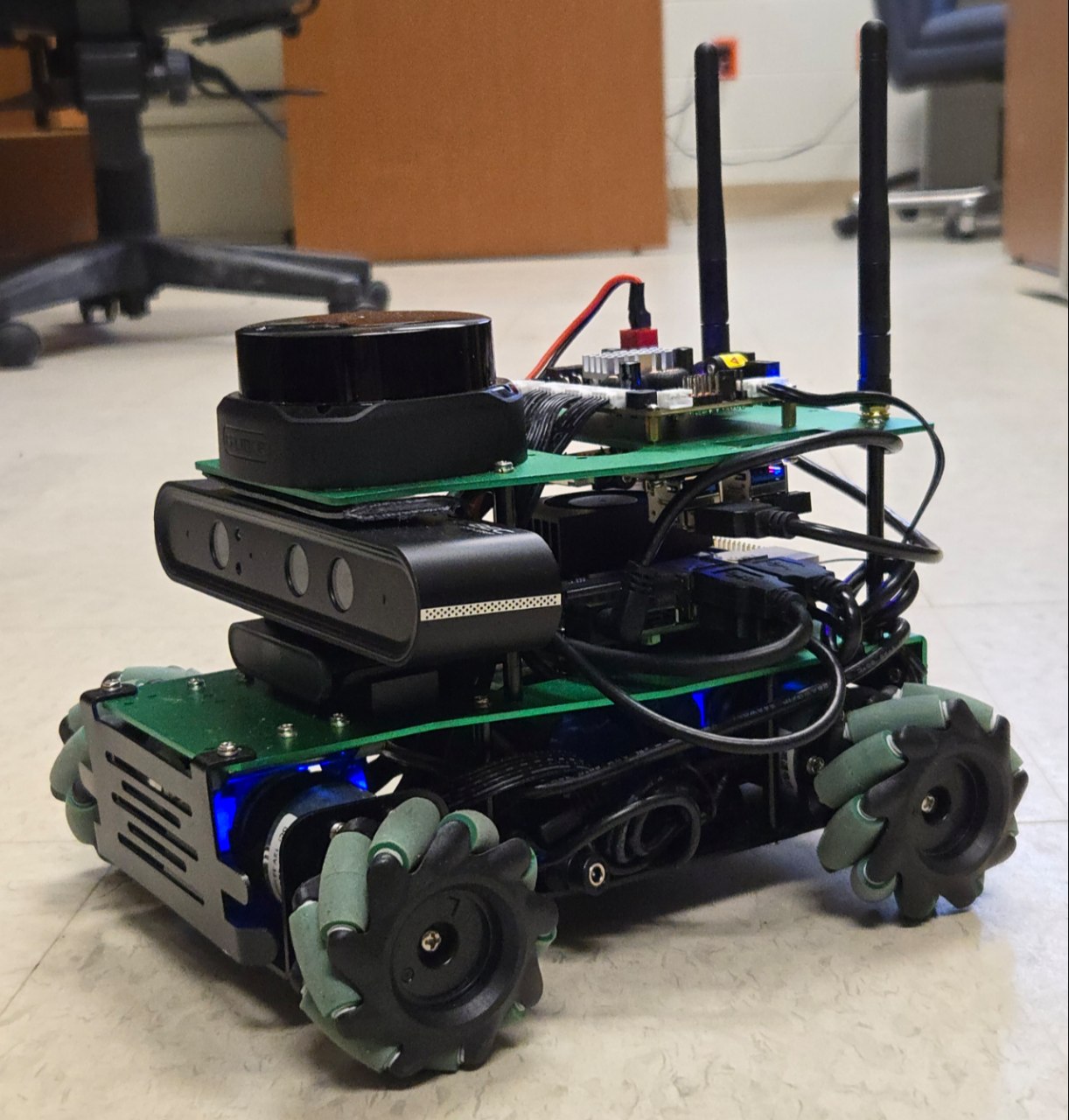}
  \caption{Robotic Platform}
  \label{fig:robot}
  \vspace{-0.75cm}
\end{figure}

During the development and experiments, two systems could communicate via WiFi, and also ROS2 core APIs and libraries provided a monitoring system over WiFi, making the process easier. As illustrated in Figure \ref{fig:robot}, our Robot is equipped with Mecanume Wheels that allow a mobile robot to have more movement types. This wheel type lets the robot move on the $x$ and $y$ axis alongside the rotational movement around the $z$ axis. However, with this wheel type, a robot can have more freedom, but modeling the movement and calculating the odometry based on the wheel's encoder would be more challenging.

\begin{table}[t!]
\centering
\caption{Robotic Platform Sensors List}
\begin{tabular}{c c c}
\toprule
\textbf{Sensor} & \textbf{Model} & \textbf{Feature} \\ 
\midrule
 2D LiDAR & \begin{tabular}{@{}c@{}} Slamtec \\ RPLIDAR S2\end{tabular}    & \begin{tabular}{@{}c@{}c@{}} Low Cost \& 360 degree \\ 
 Scanning frequency 10Hz \\ Scanning Range 0.05 - 50m \end{tabular} \\
 WiFi & Intel 8265 AC & 2.4, 5 GHz \\ 
 IMU & MPU-9250 & \begin{tabular}{@{}c@{}} Nine-Axis \\ Scanning frequency 400Hz \end{tabular} \\
 Camera & \begin{tabular}{@{}c@{}} ASTRA \\ PRO PLUS \end{tabular}  & \begin{tabular}{@{}c@{}c@{}} RGB \& Depth Camera \\ 30fps \\ Range 0.6 – 8m \end{tabular} \\
 Wheel Encoders & Hall Encoder & two hall-effect sensors \\
 \bottomrule
\end{tabular}
\vspace{0.2cm}
\label{tab:robot_info}
\vspace{-0.75cm}
\end{table}

\subsection{\textbf{Location}}
The experiment was conducted within an office in our University's Main Building. This space served as a controlled environment conducive to generating WiFi fingerprinting datasets. The room dimensions spanned 4 meters by 10 meters, yielding an area of 40 square meters. Figure \ref{fig:fingerprint} (a) illustrates a heatmap representation of the constructed WiFi fingerprinting dataset overlaid on a 2D map generated by the SLAM algorithm that shows the obstacles and the office boundaries. During the experiments, we found WiFi signals of 38 different APs in our office. Some APs are part of the university internet network, with an identical SSID, but their MAC address is different. However, these APs are usually installed in the hallways or offices, not in the line of sight. Another group of APs we found are the printers and other IoT-enabled devices placed in different offices in the building. Some APs are mobile hotspots, which might not be valid in the future online localization because they might not exist or be placed elsewhere. 

\begin{table*}[t!]
\centering
\caption{Effectiveness evaluation of the proposed approach in comparing with other approaches}
\begin{tabular}{c c c c c}
\toprule
\textbf{Method} & \begin{tabular}{@{}c@{}} \textbf{Time Efficiency} \\ (\textit{Reference Points per Second}) \end{tabular} & \textbf{Surveying Platform} & \textbf{Prepare Environments} & \textbf{Odometry Method} \\
\midrule
Ours & 1  & Robot  & No  & LiDAR SLAM  \\
\cite{abu_kharmeh_indoor_2023} & 0.14 & Robot & Yes & Black Tape\\
% \cite{kozlowski_h4lo_2020} & n/a & Bike Helmet (individuals) & No & LiDAR SLAM \\ 
\cite{rizk_laser_2023} & n/a & Crowdsourcing (individuals) & Yes & Laser-Range Scan Tracking\\ 
\cite{silva_industrial_2023} & 0.62 & Manually Pushed Trolley & Yes & ArUco tags with camera\\ 
\bottomrule
\end{tabular}
\vspace{0.2cm}
\label{table:comparison}
\vspace{-0.5cm}
\end{table*}

\begin{table}[t!]
\caption{Ground Truth Comparison}
\centering
\begin{tabular}{c c c c c}
\toprule
\textbf{Method} & \textbf{Time(seconds)} & \textbf{RP/m\(^2\)} & \textbf{RP} & \textbf{RP/s}\\
\midrule
Ours & 320 & 7.27 & 320 & 1 \\
Baseline & 656 & 0.57 & 23 & 0.035 \\
Baseline Dense & 891 & 1.17 & 47 & 0.052 \\
\bottomrule
\end{tabular}
\vspace{0.2cm}
\label{table:comparison_truth}
\vspace{-0.5cm}
\end{table}

\subsection{\textbf{Ground Truth}}
\label{label:ground_truth}
To establish our experiment's ground truth, we used a robot to collect data systematically. The data collection process was performed in two different grid densities: a coarse grid with a 0.99-meter spacing and a finer grid with a 0.66-meter spacing. In order to prevent extra work, we used the bricks on the ground, which are 0.33x0.33 meters. The Robot was navigated across a predefined grid on the ground, collecting data at each grid point. To streamline the data collection, we developed a Python application that facilitated the process via a user-friendly web interface. This web interface could be accessed over WiFi, allowing for remote control and data management. The procedure for data collection was as follows:
\begin{itemize}
    \item \textbf{Grid Navigation:} The robot was moved to a specific location on the grid.
    \item \textbf{Data Collection:} Using the web interface, the exact position of the robot was set, and the "Collect" button was clicked to initiate data collection. The robot scanned for WiFi signals at that location.
    \item \textbf{Position Logging:} The collected WiFi data and corresponding positional information were recorded.
    \item \textbf{Iteration:} The robot was then moved to the next grid point, and the process was repeated until the entire lab area was surveyed.
    \item \textbf{Exporting:} After collecting data from all possible grid points, we could save the gathered information into a single file from the web interface. 
\end{itemize}

Each WiFi scan produced data associated with a specific RP, including the Robot's location coordinates. This information was stored in a JSON format, facilitating easy access and analysis. The JSON file contained an array of objects, each representing a scan with fields for location coordinates and the corresponding WiFi scan data. By using two different grid densities, we aimed to compare the impact of grid size on the accuracy and reliability of the data collected. The finer grid was expected to provide more detailed and potentially more accurate ground truth data, while the coarser grid would offer a quicker but less detailed overview. Overall, this methodical approach to ground truth data collection ensured a comprehensive and structured dataset, crucial for the subsequent stages of our experiment.

\section{Evaluation}
\label{sec:evaluation}

In this section, we evaluate our proposed methodology from different aspects. First, the effectiveness of WiFi fingerprinting dataset creation explicitly focuses on the time efficiency and adaptability aspects; second, localization model accuracy compared to state-of-the-art works; and finally, the effect of grid density and RPs count in localization error. As shown in Table \ref{table:comparison}, we prepared a high-level summary of other similar fingerprint collection methods, which helps to compare our method and show its strengths. We compared time efficiency and adaptability factors in the following section. Then, we compared the accuracy of different localization methods concerning the RP count and technology employed, summarized in Table \ref{table:accuracy}. Finally, another comparison was made only to evaluate the effect of RP count while other parameters were the same. 

\subsection{\textbf{Time Efficiency}}
Our method outperforms existing approaches in terms of time efficiency. Abu Kharame et al. \cite{abu_kharmeh_indoor_2023} utilized a robotic platform for dataset construction, but their experiment required approximately 3 hours to collect 1500 samples using three WiFi modules. In contrast, our method achieves better speed, gathering data and building a map much faster. Our method can gather approximately one scan per second, while Abu Kharame et al.'s \cite{abu_kharmeh_indoor_2023} work speed is 0.14 scan per second. Besides the mentioned work, Silva et al. \cite{silva_industrial_2023}, which is not based on a robotic platform, require more time to collect the same amount of WiFi RPs. They could collect 0.62 RPs per second based on their experiment result. Despite the bottleneck posed by the WiFi scan rate, which is limited to one scan per second, our method still outpaces previous works considerably. Specifically, our methodology enables data collection approximately seven times faster than the previous robotic work by Abu Kharame et al. This substantial reduction in time is attributed to the adaptive nature of our approach, which eliminates the need for preparatory steps.

\begin{table*}[t!]
\centering
\caption{Comparison of different methods based on various metrics}
\begin{tabular}{c c c c c c c c}
\toprule
\textbf{Method} & \textbf{Year} & \textbf{Model} & \textbf{RMSE} & \textbf{MAE} & \textbf{RF} & \textbf{RP/m\(^2\)} & \textbf{Data} \\
\midrule
Ours & 2024 & DNN & \textbf{0.27} & \textbf{0.19} & \textbf{320} & \textbf{7.2} & RSSI \\
Rizk\cite{rizk_laser_2023} & 2023 & LSTM & - & 0.67 & 128 & 0.53 & RSSI \\
Molina\cite{molina_multimodal_2018} & 2018 & WKNN & - & $>$5 & 461 & - & RSSI \& BLE \\
Sarcevic\cite{sarcevic_indoor_2023} & 2023 & MLP & 0.51 & 0.34 & 426 & 4.43 & RSSI \& Magnometer \\
Rana\cite{rana2023indoor} & 2023 & DNN \& RF & 0.34 & - & 36 & 0.51 & RTT \\
% WiDeep & & & & 0.72 &  & 0.91 & \\
\bottomrule
\end{tabular}
\label{table:accuracy}
\vspace{-0.5cm}
\end{table*}

Table \ref{table:comparison_truth} compares our three experiments focusing on efficiency and speed. As we explained earlier in Section \ref{label:ground_truth}, two ground truth fingerprinting datasets were used to evaluate the validity of our methodology. We calculated the pairwise error of each RP in the ground truth dataset with the closest point in our approach dataset. This evaluation showed that the average error among all of the scans from ground truth and the dataset is less than 3dBm, which shows that our robotic approach can map the environment correctly. We made a comparison with the two ground truths namely \texttt{Baseline} and \texttt{Baseline Dense} provided in Table \ref{table:comparison_truth}, our robotic method collects more than six times denser datasets than a traditional map-building process and 19 times more time efficient than the traditional approaches.

\subsection{\textbf{Adaptability}}
Another significant advantage of our proposed methodology is its adaptability to various environments without requiring extensive preparation or manual work. Unlike some existing approaches (e.g., \cite{abu_kharmeh_indoor_2023, rizk_laser_2023, kozlowski_h4lo_2020, silva_industrial_2023}) that rely on predefined grids or markers or require manual work and individuals effort, our method is inherently adaptive and capable of seamlessly operating in diverse indoor settings. By leveraging a robotic platform equipped with multiple sensors and employing SLAM integration, our methodology facilitates efficient data collection and map building, irrespective of the environment's layout or characteristics. This adaptability enhances our approach's versatility and contributes to its practical applicability in real-world scenarios.

\subsection{\textbf{Localization Accuracy}}\label{sec:evaluation-accuracy}
This section evaluates and compares our proposed localization model with other established methods. We utilize multiple error metrics to provide a comprehensive assessment of model performance. Additionally, we consider the RP counts per area, the machine learning model used for localization, and the technology and the data employed by each method. The metrics used for comparison are Root Mean Square Error (RMSE) and Mean Absolute Error (MAE). Below, we provide the definitions and formulas for these metrics, followed by the comparison table.

\textbf{Root Mean Square Error}: measures the square root of the average of squared differences between actual and predicted values, providing insight into the magnitude of the error while heavily penalizing more significant errors.
\begin{equation}
\text{RMSE} = \sqrt{\frac{1}{n} \sum_{i=1}^{n} (y_i - \hat{y}_i)^2}
\end{equation}

\textbf{Mean Absolute Error}: calculates the average of the absolute differences between actual and predicted values, offering a straightforward interpretation of error magnitude.
\begin{equation}
\text{MAE} = \frac{1}{n} \sum_{i=1}^{n} |y_i - \hat{y}_i|
\end{equation}

In comparing the methods that used RSSI as their input models, our approach outperforms other methods with a considerably lower error rate. While Sarcevic et al. used Magnometer data alongside the RSSI to train their model, our approach has a lower error rate in both metrics, which is related to the higher density of our fingerprinting dataset. To show our method's superiority, we compared our result with Rana's, who has used RTT for indoor localization, which can provide more information from WiFi infrastructure. Our location prediction precision is better than their model, although we only used RSSI. This comparison shows the effect of RP counts on the localization models.

\subsection{\textbf{Reference Point Count Effect}}
\begin{figure}[t!]
  \centering
    %\includesvg[width=\linewidth]{figs/grid_effect.svg}
    \includegraphics[width=\linewidth]{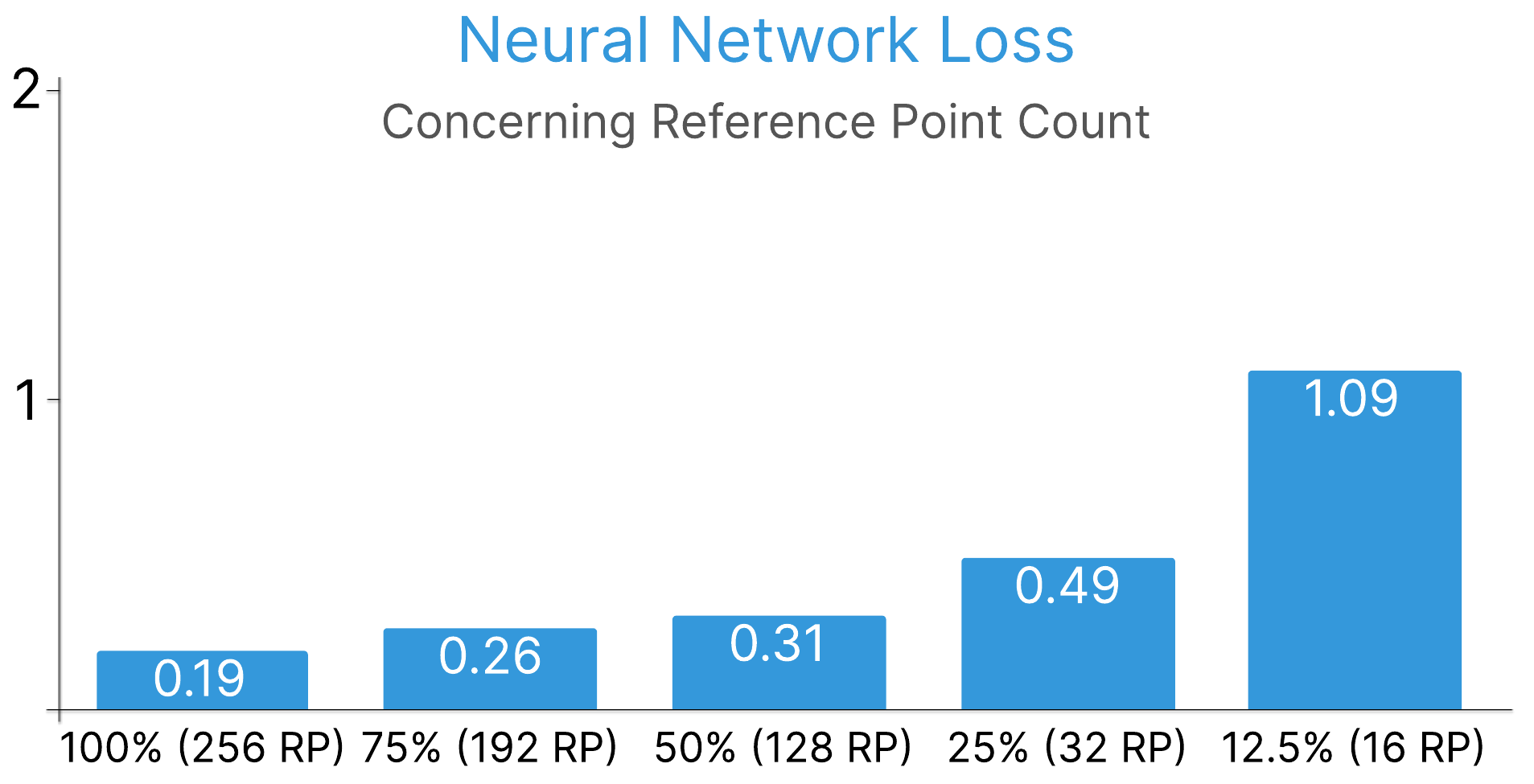}
  \caption{Reference Point Count Effect}
  \label{fig:rp_effect}
  \vspace{-0.5cm}
\end{figure}

In our final analysis, we evaluated the effect of RP counts on the localization model accuracy. Our approach can collect denser RP's grid in smaller amounts of time, which lets us train a neural network for indoor localization that outperforms state-of-the-art approaches. We conducted multiple training sessions on our deep model with different amounts of training data and created Figure \ref{fig:rp_effect} to illustrate the result. In the Figure, the $y$ axis shows the MAE error, and on the $y$ axis, we provided the training data percentage and RP count used for that training session. As we expected, with the random reduction of the training data in different training sessions, while we preserved the spatial diversity of the training samples, the model error increased considerably. If we use only half the training dataset, the model error will increase by 60\%, and by using a quarter of the fingerprinting dataset, the model error will increase by 160\%, which shows the importance of the dataset's density and training data for accurate indoor localization.

\section{Conclusion and Future Work}
\label{sec:conclusion}
We improved the fingerprinting indoor localization systems by addressing the gaps caused by inefficient and time-consuming fingerprinting dataset collection methods that can not build a dense data grid. Our proposed method offers a solution to expedite the creation of fingerprinting datasets for indoor localization. We built a fingerprinting dataset six times denser while the collection time was reduced three times. We collected a comprehensive WiFi fingerprinting dataset by applying the SLAM algorithm to real-world data for robot odometry calculation and mapping WiFi scans to various points using odometry information. This dataset serves as a valuable resource for indoor localization, accelerating the offline phase and enhancing accuracy by increasing the number of available reference points. As mentioned in the \ref{sec:sec:evaluation-accuracy}, our method achieves 26\% more accurate localization than the other methods.

In future work, our research aims to advance the capabilities of robotic platforms for building WiFi fingerprinting datasets by focusing on autonomous environment surveying. Currently, an individual must operate the robot; although controlling it with the joystick is relatively easy, it requires human intervention. We must handle challenges such as path planning, navigation, and SLAM to achieve autonomy, enabling robots to operate without human supervision in dynamic environments. Additionally, we plan to refine the mapping process to generate denser and more accurate maps using clustering algorithms and noise reduction techniques. Integration with emerging technologies, such as machine learning and sensor fusion, will further enhance the capabilities of the robotic platform. Finally, the proposed system must survey the environment in one continuous trial, which causes challenges in building a fingerprinting dataset for a multistory building. We want to utilize SLAM\_TOOLBOX features to develop our system with an update and merging abilities for vaster environment mapping. Overall, these future research directions aim to contribute to advancing robotic systems for environment surveying and mapping, ultimately facilitating various applications in different domains.

\section*{Acknowledgment}
This work was supported by the NBIF Talent Recruitment Fund (TRF2003-001). The equipment used in the experiments was supported by CFI Project Number 39473 - Smart Campus Integration and Testing (SCIT Lab).

\section*{References}
\printbibliography[heading=none]

@article{turgut_indoor_2016,
  title={Indoor localization techniques for smart building environment},
  author={Turgut, Zeynep and Aydin, Gulsum Zeynep Gurkas and Sertbas, Ahmet},
  journal={Procedia computer science},
  volume={83},
  pages={1176--1181},
  year={2016},
  publisher={Elsevier}
}

@inproceedings{rana2023indoor,
  title={Indoor Positioning using DNN and RF Method Fingerprinting-based on Calibrated Wi-Fi RTT},
  author={Rana, Lila and Dong, Jiabin and Cui, Shuyu and Li, Jinlong and Hwang, Jungyu and Park, Joongoo},
  booktitle={2023 13th International Conference on Indoor Positioning and Indoor Navigation (IPIN)},
  pages={1--6},
  year={2023},
  organization={IEEE}
}

@article{cao2020holistic,
  title={A holistic overview of anticipatory learning for the internet of moving things: research challenges and opportunities},
  author={Cao, Hung and Wachowicz, Monica},
  journal={ISPRS International Journal of Geo-Information},
  volume={9},
  number={4},
  pages={272},
  year={2020},
  publisher={MDPI}
}

@article{cao2023fostering,
  title={Fostering new vertical and horizontal IoT applications with intelligence everywhere},
  author={Cao, Hung and Wachowicz, Monica and Richard, Rene and Hsu, Ching-Hsien},
  journal={Collective Intelligence},
  volume={2},
  number={4},
  pages={26339137231208966},
  year={2023},
  publisher={SAGE Publications Sage UK: London, England}
}

@misc{mesmoudi_wireless_2013,
	title = {Wireless sensor networks localization algorithms: a comprehensive survey},
	url = {http://arxiv.org/abs/1312.4082},
	shorttitle = {Wireless sensor networks localization algorithms},
	number = {{arXiv}:1312.4082},
	publisher = {{arXiv}},
	author = {Mesmoudi, Asma and Feham, Mohammed and Labraoui, Nabila},
	date = {2013-12-14},
	eprinttype = {arxiv},
	eprint = {1312.4082 [cs]},
}

@article{shang_overview_2022,
	title = {Overview of {WiFi} fingerprinting-based indoor positioning},
	volume = {16},
	rights = {© 2022 The Authors. {IET} Communications published by John Wiley \& Sons Ltd on behalf of The Institution of Engineering and Technology},
	issn = {1751-8636},
	url = {https://onlinelibrary.wiley.com/doi/abs/10.1049/cmu2.12386},
	doi = {10.1049/cmu2.12386},
	pages = {725--733},
	number = {7},
	journaltitle = {{IET} Communications},
	author = {Shang, Shuang and Wang, Lixing},
	date = {2022},
	langid = {english},
}

@article{dai_survey_2023,
	title = {A Survey of Latest Wi-Fi Assisted Indoor Positioning on Different Principles},
	volume = {23},
	rights = {http://creativecommons.org/licenses/by/3.0/},
	issn = {1424-8220},
	url = {https://www.mdpi.com/1424-8220/23/18/7961},
	doi = {10.3390/s23187961},
	pages = {7961},
	number = {18},
	journaltitle = {Sensors},
	author = {Dai, Jihan and Wang, Maoyi and Wu, Bochun and Shen, Jiajie and Wang, Xin},
	date = {2023-01},
	langid = {english},
	note = {Number: 18
Publisher: Multidisciplinary Digital Publishing Institute},
}

@article{carotenuto_indoor_2019,
	title = {An Indoor Ultrasonic System for Autonomous 3-D Positioning},
	volume = {68},
	issn = {1557-9662},
	url = {https://ieeexplore.ieee.org/abstract/document/8457226},
	doi = {10.1109/TIM.2018.2866358},
	pages = {2507--2518},
	number = {7},
	journaltitle = {{IEEE} Transactions on Instrumentation and Measurement},
	author = {Carotenuto, Riccardo and Merenda, Massimo and Iero, Demetrio and Della Corte, Francesco G.},
	date = {2019-07},
	note = {Conference Name: {IEEE} Transactions on Instrumentation and Measurement},
}

@article{yu_novel_2019,
	title = {A Novel {NLOS} Mitigation Algorithm for {UWB} Localization in Harsh Indoor Environments},
	volume = {68},
	issn = {0018-9545, 1939-9359},
	url = {https://ieeexplore.ieee.org/document/8550815/},
	doi = {10.1109/TVT.2018.2883810},
	pages = {686--699},
	number = {1},
	journaltitle = {{IEEE} Transactions on Vehicular Technology},
	shortjournal = {{IEEE} Trans. Veh. Technol.},
	author = {Yu, Kegen and Wen, Kai and Li, Yingbing and Zhang, Shuai and Zhang, Kefei},
	date = {2019-01},
	langid = {english},
}

@inproceedings{phutcharoen_accuracy_2020,
	title = {Accuracy Study of Indoor Positioning with Bluetooth Low Energy Beacons},
	url = {https://ieeexplore.ieee.org/abstract/document/9090691},
	doi = {10.1109/ECTIDAMTNCON48261.2020.9090691},
	author = {Phutcharoen, Kanyanee and Chamchoy, Monchai and Supanakoon, Pichaya},
	date = {2020-03},
}

@article{basiri_indoor_2017,
	title = {Indoor location based services challenges, requirements and usability of current solutions},
	volume = {24},
	issn = {1574-0137},
	url = {https://www.sciencedirect.com/science/article/pii/S1574013716301782},
	doi = {10.1016/j.cosrev.2017.03.002},
	pages = {1--12},
	journaltitle = {Computer Science Review},
	shortjournal = {Computer Science Review},
	author = {Basiri, Anahid and Lohan, Elena Simona and Moore, Terry and Winstanley, Adam and Peltola, Pekka and Hill, Chris and Amirian, Pouria and Figueiredo e Silva, Pedro},
	date = {2017-05-01},
}

@article{molina_multimodal_2018,
	title = {A Multimodal Fingerprint-Based Indoor Positioning System for Airports},
	volume = {6},
	issn = {2169-3536},
	url = {http://ieeexplore.ieee.org/document/8270589/},
	doi = {10.1109/ACCESS.2018.2798918},
	pages = {10092--10106},
	journaltitle = {{IEEE} Access},
	shortjournal = {{IEEE} Access},
	author = {Molina, Benjamin and Olivares, Eneko and Palau, Carlos Enrique and Esteve, Manuel},
	urldate = {2023-12-19},
	date = {2018},
	langid = {english},
}

@inproceedings{estrada_wifi_2023,
	title = {{WiFi} Indoor Positioning System Based on {OpenWRT}},
	url = {https://ieeexplore.ieee.org/abstract/document/10199056},
	doi = {10.1109/EUROCON56442.2023.10199056},
	eventtitle = {{IEEE} {EUROCON} 2023 - 20th International Conference on Smart Technologies},
	pages = {728--733},
	booktitle = {{IEEE} {EUROCON} 2023 - 20th International Conference on Smart Technologies},
	author = {Estrada, Rebeca and Valeriano, Irving and Aizaga, Xavier and Vargas, Lourdes and Vera, Nelson and Zambrano, Diego},
	urldate = {2023-12-20},
	date = {2023-07},
}

@article{liu_survey_2020,
	title = {Survey on {WiFi}-based indoor positioning techniques},
	volume = {14},
	copyright = {© 2021 The Institution of Engineering and Technology},
	issn = {1751-8636},
	url = {https://onlinelibrary.wiley.com/doi/abs/10.1049/iet-com.2019.1059},
	doi = {10.1049/iet-com.2019.1059},
	language = {en},
	number = {9},
	urldate = {2023-12-19},
	journal = {IET Communications},
	author = {Liu, Fen and Liu, Jing and Yin, Yuqing and Wang, Wenhan and Hu, Donghai and Chen, Pengpeng and Niu, Qiang},
	year = {2020},
	pages = {1372--1383},
}

@article{sarcevic_indoor_2023,
	title = {Indoor 2D Positioning Method for Mobile Robots Based on the Fusion of {RSSI} and Magnetometer Fingerprints},
	volume = {23},
	rights = {http://creativecommons.org/licenses/by/3.0/},
	issn = {1424-8220},
	url = {https://www.mdpi.com/1424-8220/23/4/1855},
	doi = {10.3390/s23041855},
	pages = {1855},
	number = {4},
	journaltitle = {Sensors},
	author = {Sarcevic, Peter and Csik, Dominik and Odry, Akos},
	urldate = {2023-12-07},
	date = {2023-01},
	langid = {english},
	note = {Number: 4
Publisher: Multidisciplinary Digital Publishing Institute},
}

@article{abdullah_utmindualsymfi_2023,
	title = {{UTMInDualSymFi}: A Dual-Band Wi-Fi Dataset for Fingerprinting Positioning in Symmetric Indoor Environments},
	volume = {8},
	rights = {http://creativecommons.org/licenses/by/3.0/},
	issn = {2306-5729},
	url = {https://www.mdpi.com/2306-5729/8/1/14},
	doi = {10.3390/data8010014},
	shorttitle = {{UTMInDualSymFi}},
	pages = {14},
	number = {1},
	journaltitle = {Data},
	author = {Abdullah, Asim and Haris, Muhammad and Aziz, Omar Abdul and Rashid, Rozeha A. and Abdullah, Ahmad Shahidan},
	urldate = {2024-01-31},
	date = {2023-01},
	langid = {english},
	note = {Number: 1
Publisher: Multidisciplinary Digital Publishing Institute},
}

@article{abu_kharmeh_indoor_2023,
	title = {Indoor {WiFi}-Beacon Dataset Construction Using Autonomous Low-Cost Robot for 3D Location Estimation},
	volume = {13},
	rights = {http://creativecommons.org/licenses/by/3.0/},
	issn = {2076-3417},
	url = {https://www.mdpi.com/2076-3417/13/11/6768},
	doi = {10.3390/app13116768},
	pages = {6768},
	number = {11},
	journaltitle = {Applied Sciences},
	author = {Abu Kharmeh, Suleiman and Natsheh, Emad and Sulaiman, Batoul and Abuabiah, Mohammad and Tarapiah, Saed},
	urldate = {2024-01-05},
	date = {2023-01},
	langid = {english},
	note = {Number: 11
Publisher: Multidisciplinary Digital Publishing Institute},
}

@article{rizk_laser_2023,
	title = {Laser Range Scanners for Enabling Zero-overhead {WiFi}-based Indoor Localization System},
	volume = {9},
	issn = {2374-0353},
	url = {https://dl.acm.org/doi/10.1145/3539659},
	doi = {10.1145/3539659},
	pages = {4:1--4:25},
	number = {1},
	journaltitle = {{ACM} Transactions on Spatial Algorithms and Systems},
	shortjournal = {{ACM} Trans. Spatial Algorithms Syst.},
	author = {Rizk, Hamada and Yamaguchi, Hirozumi and Youssef, Moustafa and Higashino, Teruo},
	urldate = {2024-01-09},
	date = {2023-01-12},
}

@article{silva_industrial_2023,
	title = {Industrial Environment Multi-Sensor Dataset for Vehicle Indoor Tracking with Wi-Fi, Inertial and Odometry Data},
	volume = {8},
	rights = {http://creativecommons.org/licenses/by/3.0/},
	issn = {2306-5729},
	url = {https://www.mdpi.com/2306-5729/8/10/157},
	doi = {10.3390/data8100157},
	pages = {157},
	number = {10},
	journaltitle = {Data},
	author = {Silva, Ivo and Pendão, Cristiano and Torres-Sospedra, Joaquín and Moreira, Adriano},
	urldate = {2024-01-31},
	date = {2023-10},
	langid = {english},
	note = {Number: 10
Publisher: Multidisciplinary Digital Publishing Institute},
}

@article{kozlowski_h4lo_2020,
	title = {H4LO: automation platform for efficient {RF} fingerprinting using {SLAM}-derived map and poses},
	volume = {14},
	rights = {© 2020 The Institution of Engineering and Technology},
	issn = {1751-8792},
	url = {https://onlinelibrary.wiley.com/doi/abs/10.1049/iet-rsn.2019.0369},
	doi = {10.1049/iet-rsn.2019.0369},
	shorttitle = {H4LO},
	pages = {694--699},
	number = {5},
	journaltitle = {{IET} Radar, Sonar \& Navigation},
	author = {Kozłowski, Michał and Twomey, Niall and Byrne, Dallan and Pope, James and Santos-Rodríguez, Raúl and Piechocki, Robert J.},
	urldate = {2024-02-01},
	date = {2020},
	langid = {english},
}

@article{ros2_2022,
    author = {Steven Macenski  and Tully Foote  and Brian Gerkey  and Chris Lalancette  and William Woodall },
    title = {Robot Operating System 2: Design, architecture, and uses in the wild},
    journal = {Science Robotics},
    volume = {7},
    number = {66},
    pages = {eabm6074},
    year = {2022},
    doi = {10.1126/scirobotics.abm6074},
    URL = {https://www.science.org/doi/abs/10.1126/scirobotics.abm6074}
}

@online{yahboom,
	title = {{ROSMASTER} X3 {ROS}2 Robot with Mecanum Wheel for Jetson {NANO} 4GB/Orin {NANO}/Orin {NX}/{RaspberryPi} 4B},
	url = {https://category.yahboom.net/products/rosmaster-x3},
	titleaddon = {Yahboom},
	urldate = {2024-02-02},
	langid = {english},
}

@article{macenski_slam_2021,
	title = {{SLAM} {Toolbox}: {SLAM} for the dynamic world},
	volume = {6},
	copyright = {http://creativecommons.org/licenses/by/4.0/},
	issn = {2475-9066},
	shorttitle = {{SLAM} {Toolbox}},
	url = {https://joss.theoj.org/papers/10.21105/joss.02783},
	doi = {10.21105/joss.02783},
	language = {en},
	number = {61},
	urldate = {2024-06-07},
	journal = {Journal of Open Source Software},
	author = {Macenski, Steve and Jambrecic, Ivona},
	month = may,
	year = {2021},
	pages = {2783},
}

\end{document}